\documentclass[letterpaper, 10 pt, conference]{ieeeconf}  

\IEEEoverridecommandlockouts                              

\usepackage{graphics} 
\usepackage[dvipdfmx]{graphicx}

\usepackage{epsfig} 
\usepackage{mathptmx} 
\usepackage{times} 
\usepackage{amsmath} 
\usepackage{amssymb}  
\usepackage{cite}
\usepackage[T1]{fontenc}

\usepackage[subrefformat=parens]{subcaption}
\usepackage{soul}

\title{\LARGE \bf
Probabilistic Slide-support Manipulation Planning in Clutter
}

\author{Shusei Nagato$^{1}$, Tomohiro Motoda$^{1}$, Takao Nishi$^{1}$, Petit Damien$^{1}$,\\
Takuya Kiyokawa$^{1}$, Weiwei Wan$^{1}$, and Kensuke Harada$^{1,2}$
\thanks{$^{1}$ Shusei Nagato, Tomohiro Motoda, Takao Nishi, Petit Damien and Weiwei Wan are with the Graduate School of Engineering Science, Osaka University, Toyonaka, Osaka 560-0043, Japan}
\thanks{$^{1,2}$ Kensuke Harada is with Graduate School of Engineering Science, Osaka University, Toyonaka, Osaka 560-0043, Japan, and also with the National Institute of Advanced Industrial Science and Technology (AIST), Tokyo, Japan
}}

\begin{document}

\maketitle
\thispagestyle{empty}
\pagestyle{empty}

\begin{abstract}

To safely and efficiently extract an object from the clutter, this paper presents a bimanual manipulation planner in which one hand of the robot is used to slide the target object out of the clutter while the other hand is used to support the surrounding objects to prevent the clutter from collapsing. Our method uses a neural network to predict the physical phenomena of the clutter when the target object is moved. We generate the most efficient action based on the Monte Carlo tree search.The grasping and sliding actions are planned to minimize the number of motion sequences to pick the target object. In addition, the object to be supported is determined to minimize the position change of surrounding objects. Experiments with a real bimanual robot confirmed that the robot could retrieve the target object, reducing the total number of motion sequences and improving safety. 

\end{abstract}

\section{INTRODUCTION}
    Automation in a logistic warehouse is more and more expected following the increase in the use of e-commerce. One of the typical tasks for a robot in a logistic warehouse is to pick an item from storage and put it into a container. 
    The difficulty of picking an item from storage comes from several aspects. Since the items are densely arranged in the storage for efficiency, a robot may struggle to find an appropriate surface on the target item suitable for its gripper to realize a successful pick. In addition, even if a robot finds a grasping pose of the target item, the robot may not be able to safely pick it up since it may contact the surrounding items, resulting in a disorderly display, or the item may fall and be damaged. 
    %
    %
    There have been several attempts for a robot to pick an item from clutter. Studies focusing on this problem can be mainly classified into two groups. In the first group, a robot pushes the items to the side until it can pick the target item~\cite{c7,c8, c9, c10}. 
    On the other hand, in the second group, a robot picks the stacked items from the top until it can pick the target item. However, the above manipulation strategies are inefficient since a robot has to repeatedly perform the pushing or picking operations several times according to the target item position in the clutter. 
     

    \begin{figure}[t]
        \begin{center}
            \includegraphics[width=\linewidth]{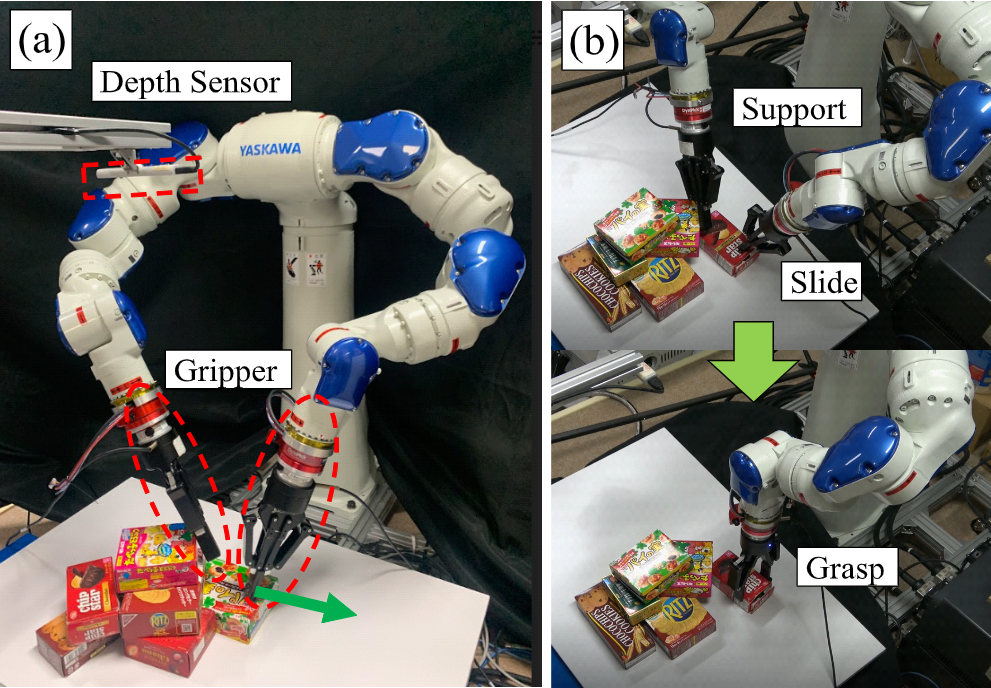} 
            \caption {Overview of our proposed method on picking the target object. 
            (a) Cluttered scene experiment setup with the robot and parallel gripper (inside the red frames); The robot slides the box from the clutter with one gripper while supporting the red boxes with its other gripper. (b) Picking a chocolate box from the slide and support. }
            \label{bimanual}
        \end{center}
    \end{figure}
    
    To cope with such a problem, this paper proposes a bimanual manipulation planner for extracting an object from clutter. 
    Our manipulation planner can efficiently minimize the number of motion sequences. To grasp the object placed at the bottom of the clutter, one hand of the robot is used to slide the target object out of the clutter, while the other hand is used to support surrounding objects to prevent the clutter from collapsing and the product from being damaged. 
    On the other hand, to grasp an object placed near the top of the pile, a robot first tries to remove the object placed on top of the target object to be able to grasp it. 
  
    The proposed planner predicts the stacking states from the observed scene of the clutter after the operations performed by the robot~(Fig.~\ref{bimanual}). 
    The transition of the stacking states is estimated using a neural network, which is trained with data of the pose of each object 
    obtained through physics simulations and trials in the real environment. 
    Our planner selects actions from grasping and sliding under the reward of minimizing both the number of motion sequences and the position change of surrounding objects during the sliding motion. Based on the observation results, the robot executes the determined movements and re-estimates the stack. 
    The observation and action planning process is then repeated several times to efficiently remove objects from the stack.

    The structure of this paper is as follows. Section~\ref{sec:related_work} presents the related work. Section~\ref{sec:main} explains our action planner to select an object safely and efficiently. In Section~\ref{sec:experiment}, we present our experimental results with a real manipulator and discuss the result and the problems in the real world. Section~\ref{sec:conclusion} concludes the paper and discusses future work.   

    \begin{figure*}[t]
        \centering
        \includegraphics[width=\linewidth]{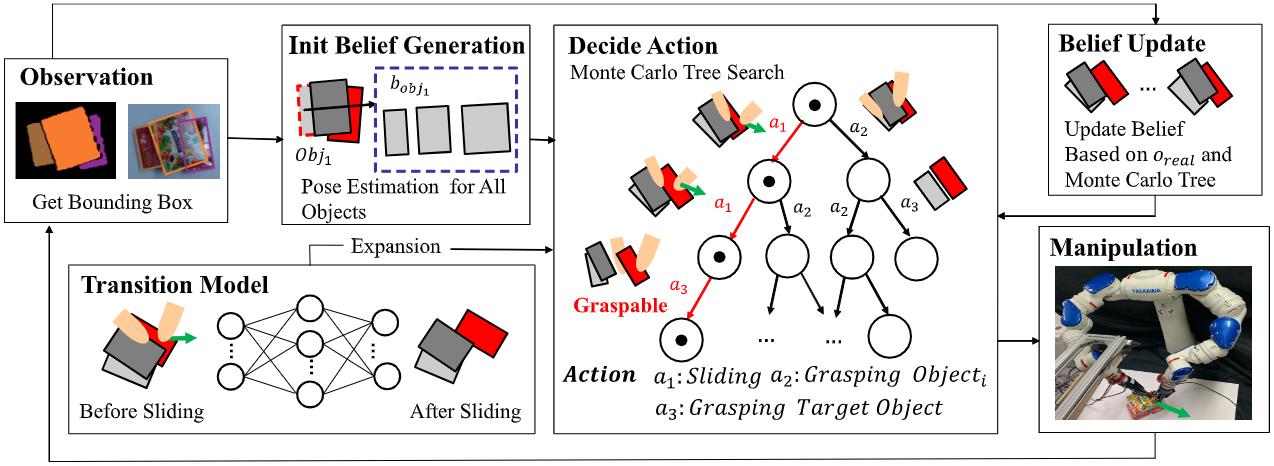} 
        \caption{Overview of our proposed method. Our proposed algorithm uses an object-based transition model to predict the motion of all objects in clutter based on a neural network, and object states are updated under new observations. The support or slide action is selected at each step to minimize the translation of non-target objects. }
        \label{overview}
    \end{figure*}

\section{RELATED WORK}
\label{sec:related_work}
    We first introduce previous works on removing obstacles to retrieve a target object in clutter~\cite{c11,c15,c16}. Ahn et al.~\cite{c14} rearranged cluttered objects with graph searches to retrieve the target object. 
    Some recent works on learning-based picking also estimate the spatial relationship of objects in clutter~\cite{c18,c19}. Motoda et al.~\cite{c27} designed a neural network to estimate the support relation of objects in clutter and predicted the clutter to collapse. However, these approaches must pick the surrounding objects several times to retrieve the target.  
    On the other hand, this paper proposes a new manipulation planner to efficiently slide the target object out of the clutter while the other hand supports the surrounding objects to prevent the clutter from collapsing. 

    Research has recently been conducted on object rearrangement based on the estimated pose of objects in the clutter. 
    Berscheid et al.~\cite{c20} planned to pick an object in clutter by rearranging the surrounding ones, assuming enough space to rearrange. Huang et al.~\cite{c22} designed a predictive neural network for pushing operations to rearrange objects. The reinforcement learning methods~\cite{c23,c24} also realized the pushing-grasping manipulation in clutter to extract the target object. Wada et al.~\cite{c25} realized the safe picking by minimizing the destructive effects on the surrounding objects. 
    However, all the above researches assumed that the pose of each object can be detected. On the other hand, our research generates sliding actions in an uncertain environment by integrating the learning-based predictive model into a partially observable Markov decision process (POMDP). 

\section{Motion Planning for Picking the Target Object In Clutter}
\label{sec:main}


\subsection{Problem Description}

    In our framework, a robot can switch to multiple manipulation strategies to extract the target object from clutter. 
    In the manipulation strategy based on sliding,    
    a bimanual robot iteratively slides the target object until it is safely extracted. 
    While one hand of the robot is used to slide the target object, the other hand is used to support a surrounding object to prevent the clutter from collapsing. 
    In the manipulation strategy based on removing objects, a robot removes the object placed on top of the target one until the robot can grasp the target one. 
    
    We assume that the clutter is composed of $N+1$ rectangular objects placed on a flat table where the target one is with a known size and the surrounding ones are with an unknown size. We also assume that each object makes facial contact with another object or the table. 
    We also assume a bimanual robot with a parallel gripper at each arm's tip. 
    The robot closes the gripper and presses the fingertip to the top surface of the target object to slide it. On the other hand, the robot closes the gripper and contacts the fingertip to the side surface of a surrounding object to support it. We use a depth sensor directing vertically downward direction to observe the clutter. 

    
    Fig.~\ref{overview} shows the overview of our proposed method. 
    First, in the observation step, we obtain the depth images of the clutter and extract bounding boxes (BBOXs) that contain their center position, orientation, and size. 
    Second, we generate the belief state of each object maintaining the uncertainty. 
    Then, the state of the manipulated object is estimated based on the transition model. 
    Third, we determine the motion sequence using the tree traversal method, Monte Carlo tree search~\cite{c32}. The actions are planned to minimize the steps of extracting the target object. 
    The details of the method are described in the following sections. 



    \begin{figure}[tb]
        \begin{center}
            \includegraphics[width =0.99\linewidth]{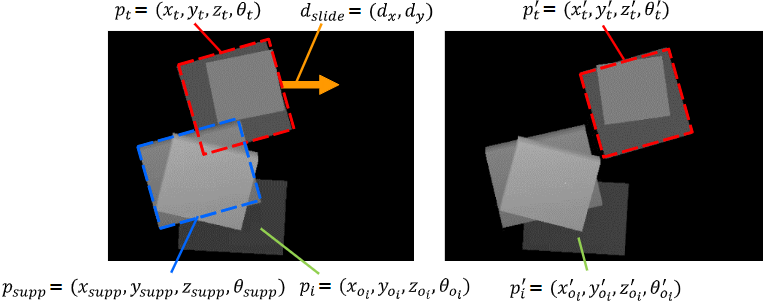} 
            \caption{Example of object states and slide actions. }
        \end{center}
        \label{record}
    \end{figure}
    
    The problem of object extraction considered in this research is formulated as a partially observable Markov decision process (POMDP) problem~\cite{c29}. The POMDP consists of a tuple $\{S,A,T,O,Z,R\}$, where $S$, $A$, $T$, $O$, $Z$ and $R$ denote the state, action, state transition model, observation, observation model and reward, respectively. 
    The POMDP model can be defined as follows: 
    \\\textbf{State:} 
    Let $S_t$ and $S_{o_i}$ ($i=1,2,...,N$) be the state of the target object and surrounding ones, respectively. 
    The state $S$ can be represented by $s = (s_t, s_{o_1}, ..., s_{o_N})$. Here, $s_t$ is composed of the position and orientation about the vertical axis of the target object, represented by $s_t = (x_t, y_t, z_t, \theta_t)$. On the other hand, $s_{o_i}$ is composed of the position, orientation about the vertical axis and the edge length of the surrounding object, represented by $s_{o_i} = (x_{o_i}, y_{o_i}, z_{o_i}, \theta_{o_i}, w_{o_i}, h_{o_i})$. Fig.~\ref{record} shows an example of the object state. 
    \\\textbf{Action:} 
    The action $A$ comprises three elements, i.e., sliding, removing, and grasping. In the sliding action, one hand slides the target object at a certain distance while the other hand supports a surrounding object. In the removing action, a hand grasps the $i$~th surrounding object and removes it from the workspace. In the grasping action, a hand grasps the target object and terminates the action. Regarding the sliding action, there are $N$ candidates for supporting an object, and we assumed 16 sliding directions within the horizontal plane.
    \\\textbf{Transition Model:} 
    Transition model $T$ represents the change of the state from $s$ to $s'$ due to an element of action $a$. The transition to the next state after a sliding action is estimated using the neural network. Similarly, for removing actions, the neural network estimates the transition from the current state to the state where the object to be grasped is removed.
    \\\textbf{Observations:}
     The observation $O=(o_{t}, o_{o_1}, ..., o_{o_N})$ includes the observed position, orientation, and size of the oriented bounding box (BBOX) of each object. The position is represented by a fixed-length grid on the workspace, while the orientation about the vertical axis is represented discretely by twelve directions.
     \\\textbf{Observation Model:}
     $Z$ denotes the observation $o^{\prime}$ in the state $s^{\prime}$. The observation $o^{\prime}$ is obtained by taking the state $s^{\prime}$ as input and estimating the BBOX of each object that is not occluded.
    \\\textbf{Reward:} 
    $R$ denotes the reward for performing the action $a$ in the state $s$ and transitioning to the next state $s^{\prime}$. 
    The reward for the sliding action is the sum of the following rewards: $R_{occ}\in[-1,0]$ which is the rate of change $r_{occ}$ in the occlusion of the target object, $R_{t}\in[-1,0]$ which is the penalty based on the total translation of surrounding objects during the slide of the target objects. 

\begin{figure}[tb]
        \centering
        \includegraphics[width=\linewidth]{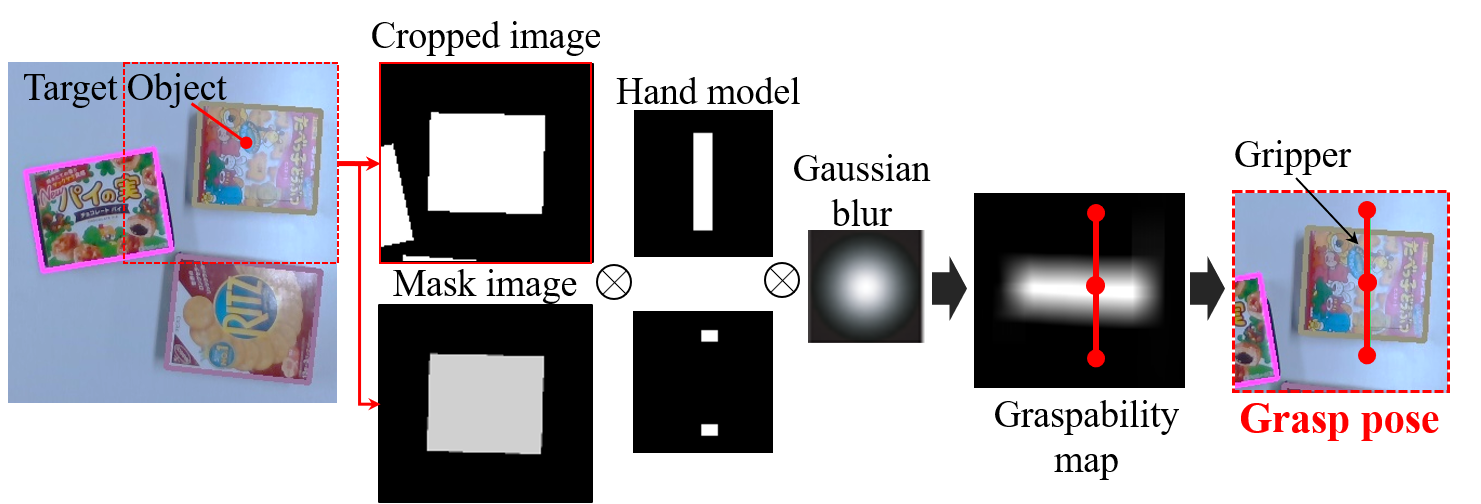}
        \caption{Grasp detection with the fast graspability evaluation~\cite{c30}. }
        \label{graspability}
    \end{figure}

    For the removing action, the reward $R_g$ is set to $-10$ for failure and $0$ for success. For grasping the target object, the reward is $-10$ for failure and $+10$ for success. The success or failure of the removing and grasping actions is determined by checking the graspability~\cite{c30} of the object to be grasped and the occlusion ratio. Graspability index evaluates the existence of the grasping pose by checking if the fingers can be placed beside the object and can grasp it, as shown in Fig.~\ref{graspability}. If the graspability is above the threshold, we consider it a successful case. On the other hand, if the occlusion ratio is large, the grasp is considered to be unsuccessful because it may result in a collapse of the clutter. 

\subsection{Prior estimation of clutter}
\label{subsec:pre_estimation}
\begin{figure}[tb]
        \begin{center}
            \includegraphics[width=\linewidth]{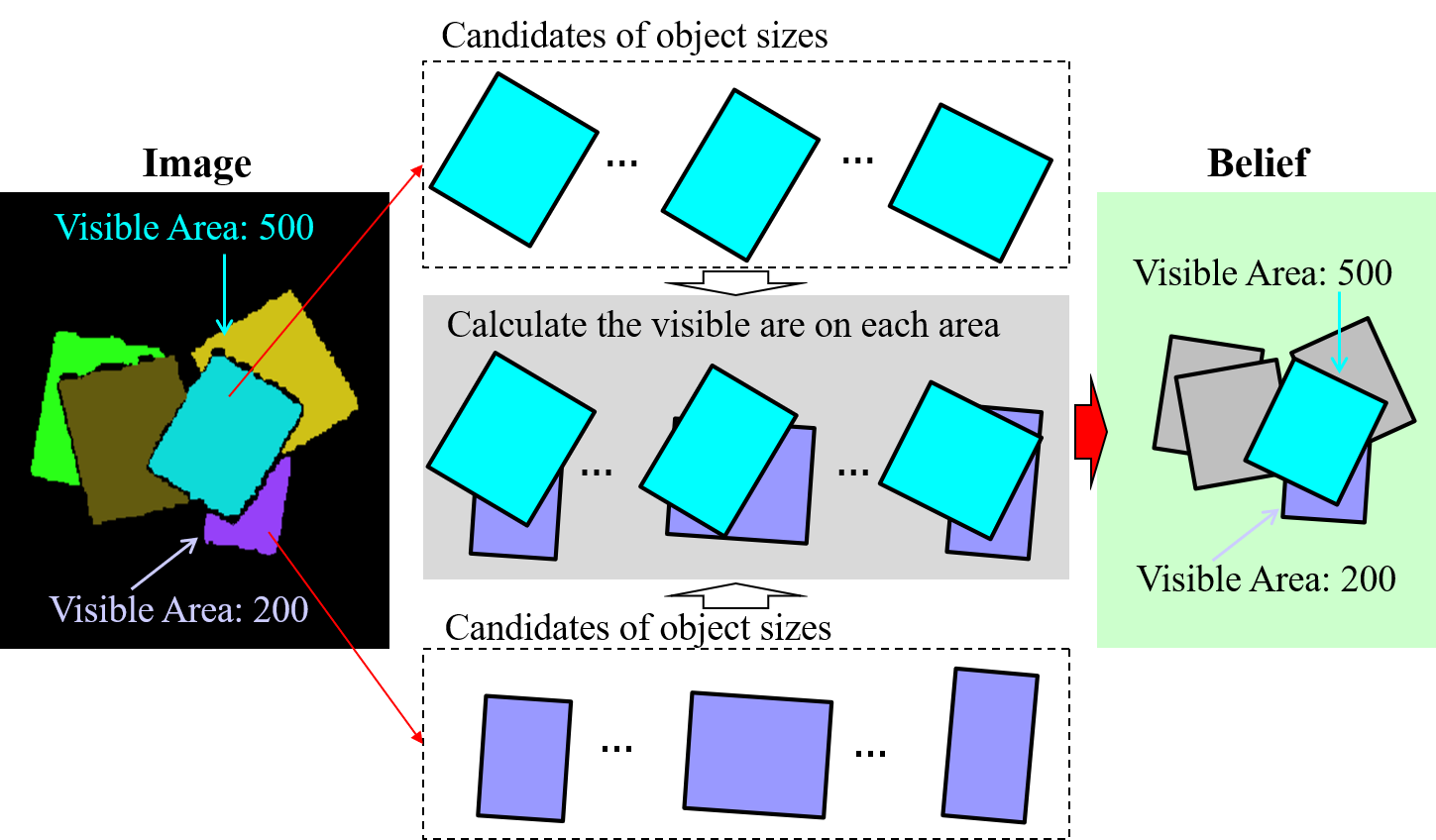}
        \caption{Estimation of the belief state. We adjust the size of each BBOX from the candidates, and select a belief state to maintain the visible areas. }
        \label{belief}
        \end{center}
    \end{figure}

    This section describes the overall process of the prior estimation. 
    Due to the occlusion, multiple patterns of states can be estimated based on the observations, and a set of estimated states have the belief. 
    To obtain the states' belief, the method shown in Fig.~\ref{belief} is used. First, the obtained depth image is segmented by using the Region Growing method~\cite{c31}. Then, the BBOXs of the segmented areas are sorted by their vertical height, and the visible area of the specified BBOX is enlarged by a certain amount. 
    If the change of the visible area after expansion is smaller than a threshold, the object is considered to exist. This process is repeated for all BBOXs until the change of the visible area after expansion is bigger than the threshold value. The possible states are calculated for each of the N objects. Then, the possible states of the N objects in all patterns are combined to create a set of hypotheses, which is the belief. 

\subsection{Object-based transition model}
    We present an object-based transition model to predict the state of objects in clutter. 

    \begin{figure}[tb]
        \centering
        \includegraphics[width =0.95\linewidth]{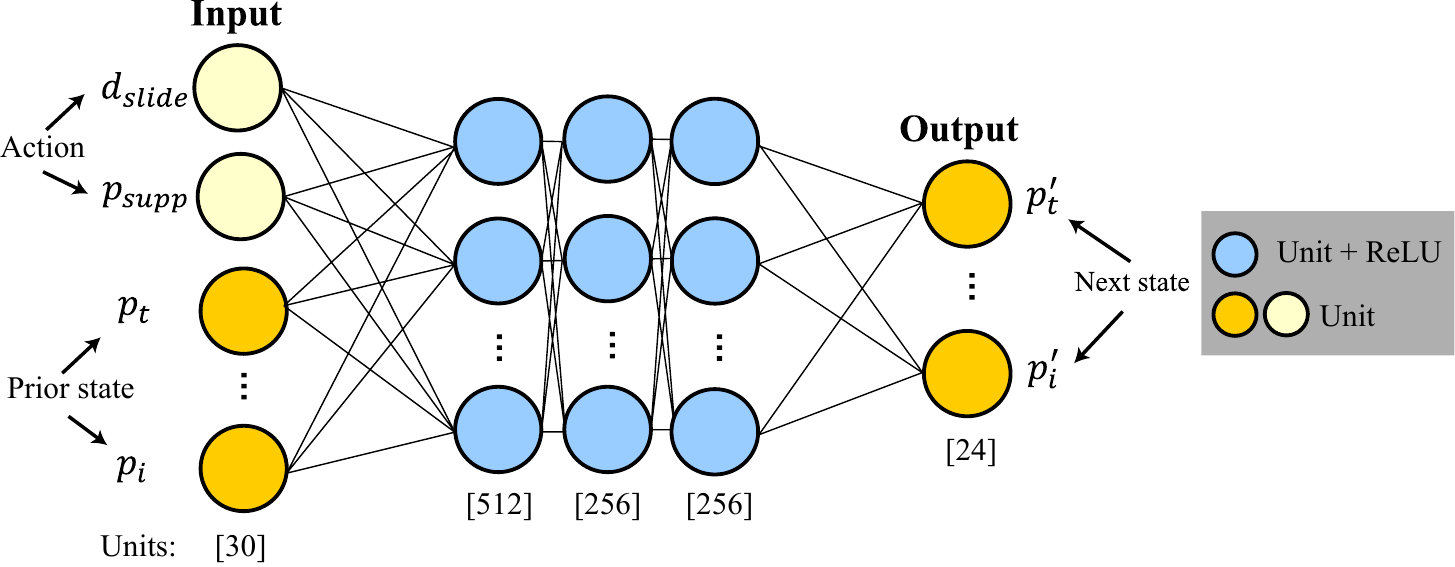}
        \caption{Architecture of our transition model based on the five-layered perceptron neural network. }
    \label{fig:neural_network}
    \end{figure}

    \begin{figure}[tb]
        \begin{center}
            \includegraphics[width=\linewidth]{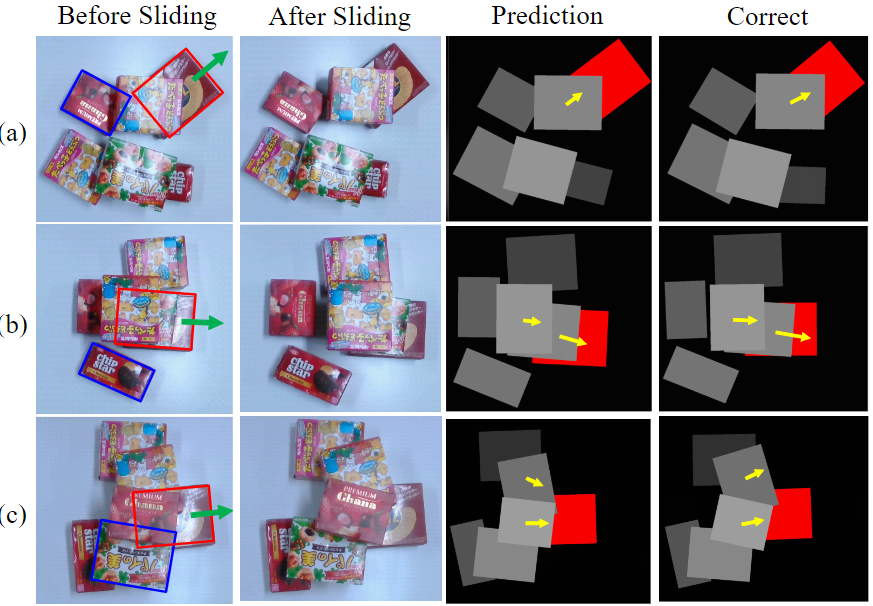}
            \caption{Examples of prediction results. For (a-c), the left two columns show the real scenes before and after the target box (red) is slid in the direction of the green arrow. The right two columns are the simulated and actual sliding images before and after sliding the target box (red) in the direction of the green arrow. The images in the two right columns also show yellow arrows of direction and length corresponding to the direction and amount of movement of the target before and after sliding. }
            \label{fig:prediction}
        \end{center}
    \end{figure}

    \subsubsection {Network Architecture}
    A Five-layered perceptron neural network is used to train the transition model, as shown in Fig.~\ref{fig:neural_network}. 
    Our model is designed to predict the posterior state $s'$ ($4(N+1)$ dimensions) using the prior state $s$ ($4(N+1)$ dimensions) and the sliding action $a$ ($6$ dimensions). 
    This network is consisted of three hidden layers with 512, 256 and 256 units. 
    Each hidden layer is transformed by the rectified linear unit activation function. The training loss is the mean squared error.   
    It should be noted that this error may become large because of the object properties, like the fiction coefficient and object size. Therefore, a Gaussian distribution is added as the variance into the translation. 
    Fig.~\ref{fig:prediction} shows the experimental results of correction by using real images. 

    \subsubsection {Dataset}

        \begin{figure}[tb]
            \begin{center}
            \includegraphics[width =0.95\linewidth]{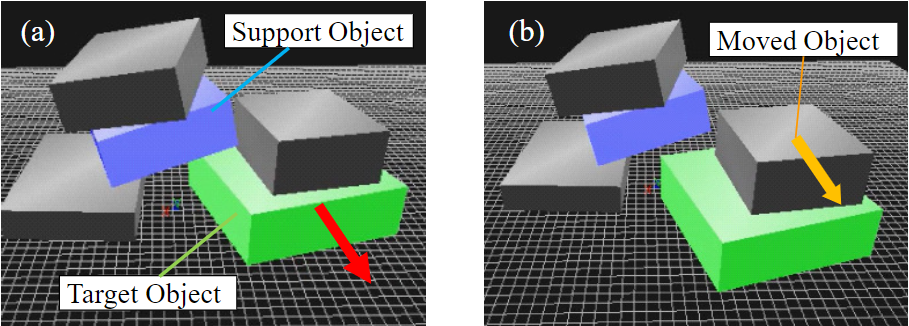} 
            \caption{Data collection with a simulator. (a) The green object moves in the direction of the red arrow while an object marked in blue remains is supported. (b) The moved object is recorded. }
            \label{sim_process}
            \end{center}
        \end{figure}
        
        The transition model is trained with a dataset obtained both from physics simulation and real experiments. 
        We can correctly predict the state of clutter if we obtain all the training data from real experiments. 
        However, it is burdensome to obtain all the training data from real experiments. Therefore, the training data obtained from physics simulation using NVIDIA PhysX is additionally used. 
        As shown in Fig.~\ref{sim_process}, several rectangular objects of various sizes are randomly placed to generate clutters.  an object is moved with a certain distance in twelve horizontal directions. At the same time, one of the objects is fixed to simulate the supported one. 
        Both the states of all the objects before and after performing an action are recorded. 
        
        To collect training data from real experiment, we place $N+1$ objects horizontally on the workspace, and segment the depth image. We identify the pose of each object by fitting a rectangular with a known size to the BBOX of each segment.  
        After obtaining the the initial pose of each object, one object randomly selected from the clutter is slid in an arbitrary direction while another object is supported. Then, we obtain the depth image after the slide motion, and detect the pose of all objects. 
        
\subsection{Motion Planning with POMCP}

    In this section, the motion planning for extracting the target object while supporting a surrounding object is presented. Our algorithm is based on the Partially Observable Monte Carlo Planning (POMCP)~\cite{c33} using Monte Carlo tree traversal. 
    Each node~$n$ in the Monte Carlo tree has a number of visits $N(n)$, a history of actions $h$, and an action value function $Q(n, a)$ for each action, and the root node has a belief $h$ in addition to these. First, one state is sampled from the beliefs. Then, the action selection is then repeated according to the Upper Confidence Bounds
    for Trees (UCT)~\cite{c34} until a node that has never been reached is reached from the current node $n_c$, and then the next state and reward according to the state transition prediction are obtained. 
    The evaluation function is expressed by the following equation.
    \begin{equation}
        \label{eq:synergy_matrix}
        UCT(n_c) = \underset{a\in A}{argmax}(Q(n_c,a) + C\sqrt{\frac{InN(n_c)}{N(N_c,a)}})
    \end{equation}
     $C$ is a search constant.
    When a new node is reached, the following rules are used to randomly select actions up to a predetermined depth.
     \begin{itemize}
      \item The robot slides the target object in the direction where the occlusion of the target object can be reduced.
      \item After the removing action, either removing or grasping actions is performed.
      \item The grasping action is performed anytime when the target object is graspable. 
     \end{itemize}
    
    The total reward $R$ obtained by these methods is used to update the action value functions of the nodes that have been traversed. 
    This process is repeated for a predetermined number of times $N_{max}$, and the robot performs the action with the highest value at the root node.
    
    POMCP used in this method treats beliefs as a set of particles representing the states, and updates beliefs using a particle filter. 
    After an action is executed, the observation is performed again, and the actual observed value $o_{real}$ is obtained. For every state that passes a node beyond the edge of the executed action, the predicted observed value $o_{pred}$ is obtained according to the observation model. The error in the Euclidean distance of the center position of the BBOX is calculated for both $o_{real}$ and all $o_{pred}$.
    The next belief is updated when the error is less than or equal to the threshold $error_{thresh}$ for all objects. If there is a large error between the prediction and the actual observation, and no state satisfying this error exists, the belief is estimated again from the BBOX using the method described in Section ~\ref{subsec:pre_estimation}. After the estimation, object retrieval is achieved by repeating the action planning and observation process using POMCP.

\section{Experiments}
\label{sec:experiment}
    
    In this section, we show how the proposed method works and analyze its performance in the real environment. 

    \subsection{Data Collection} 

    \begin{figure}[tb]
        \begin{center}
            \includegraphics[width =0.99\linewidth]{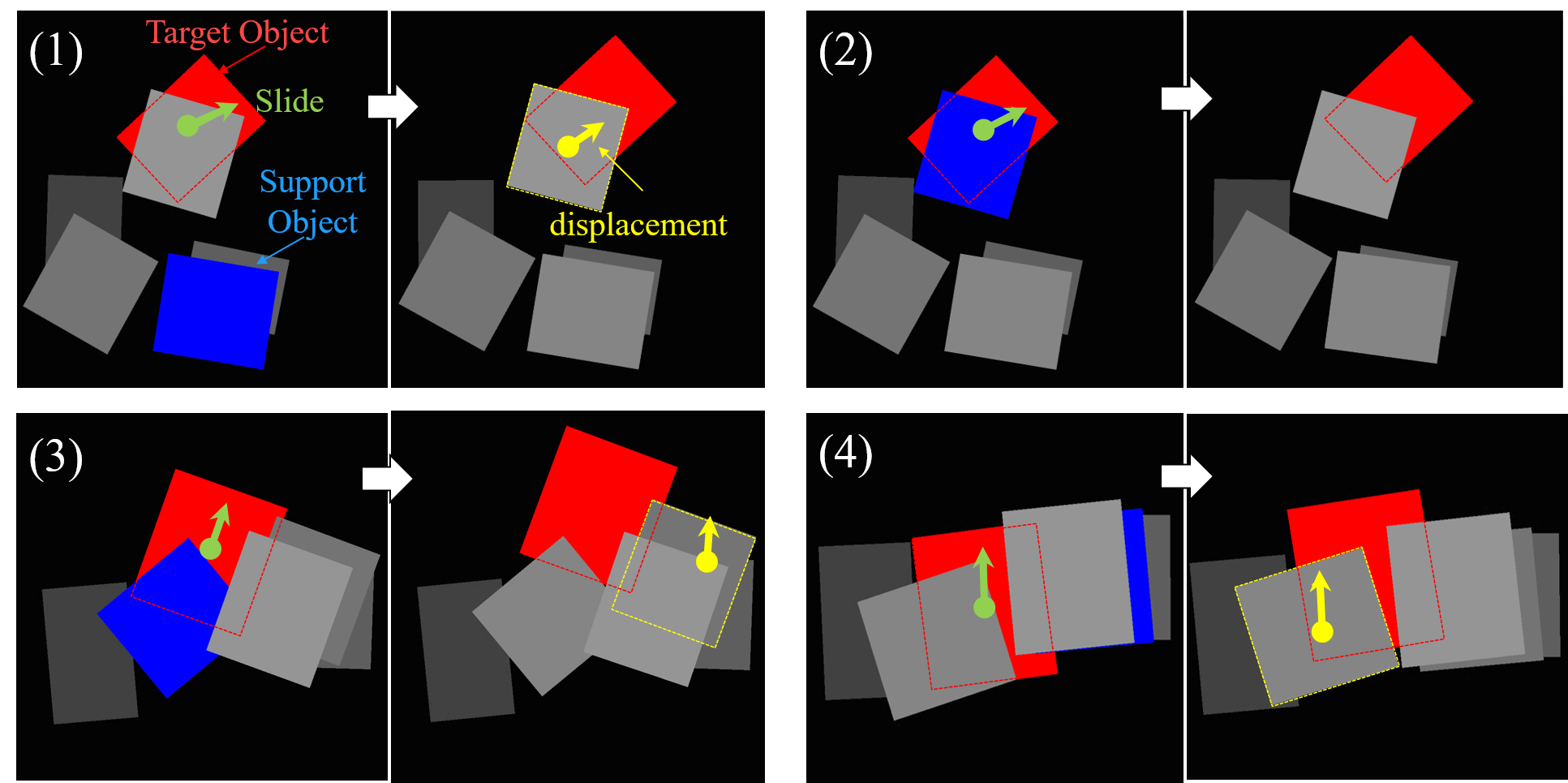}
            \caption{The visualization of the training data where the target and supported objects are marked in red and blue, respectively. }
        \label{fig:training_data}
        \end{center}
    \end{figure}

    Some examples of visualization of the collected data are shown in Fig.~\ref{fig:training_data}. In the figure, the target and supported objects are marked in red and blue, respectively, whereas the attached arrow shows the vector of object displacement. We obtained 3000 data from physics simulation. We also obtained 100 data from the real experiment where the training data is expanded to 1000 by rotating and translating the obtained images. The learning rate, batch size, and the number of epochs of the neural network were set as 0.0001, 512, and 400, respectively. 
    Adam~\cite{Adam} is used as an optimizer. 
    
    \subsection {Experimental Settings}

    \begin{figure}[t]
        \centering
        \includegraphics[width=0.95\linewidth]{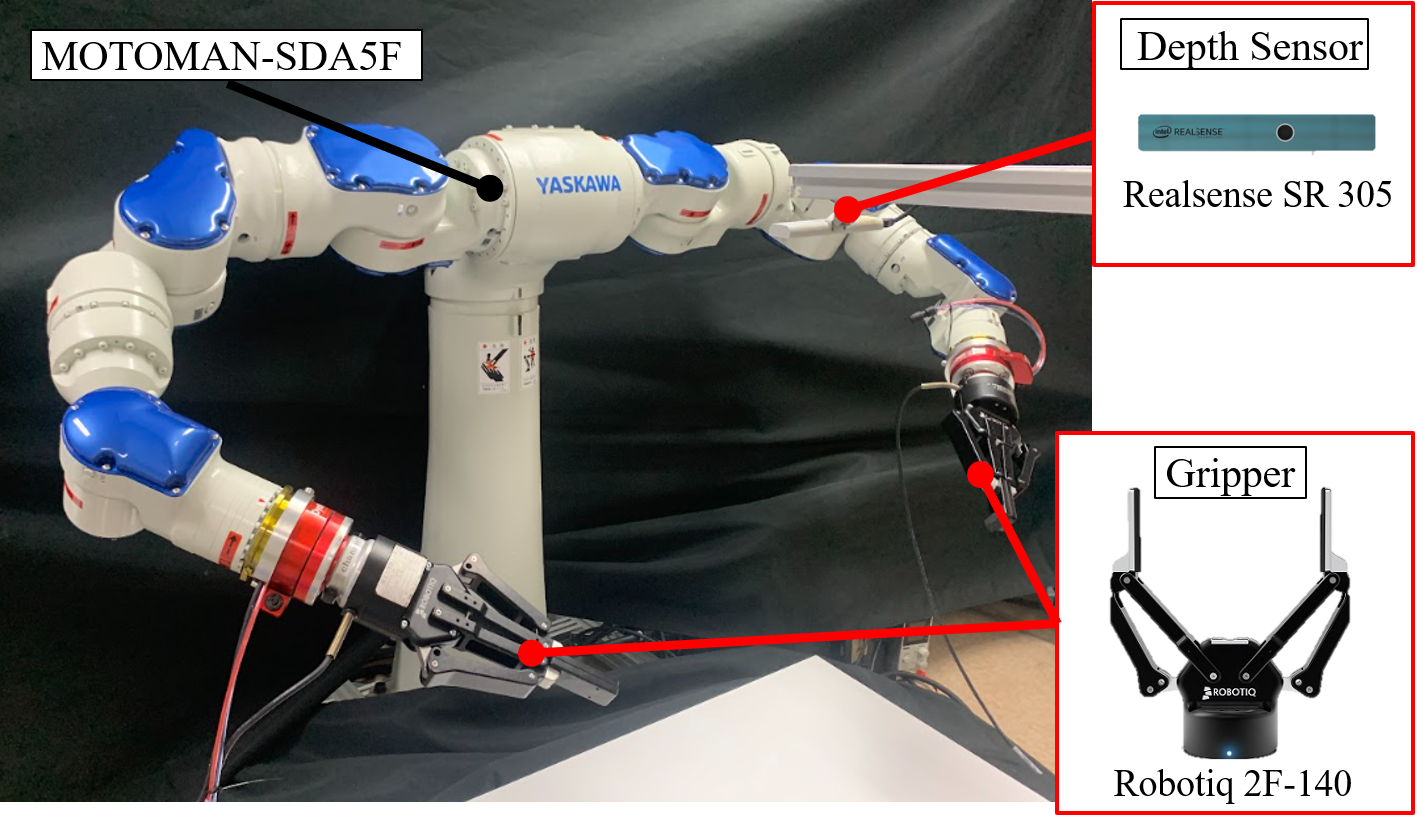}
        \caption{The robot system used in the real environment. }
        \label{env}
    \end{figure}

    The MOTOMAN-SDA5F bimanual robot, which includes two ROBOTIQ 2F-140 adaptive grippers, was used for the experiments. 
   A RealSense SR305 depth sensor was mounted on the workspace to obtain the depth images of the objects. Fig.~\ref{env} shows the robot system. 

    \begin{figure}[t]
        \begin{center}
            \includegraphics[width=\linewidth]{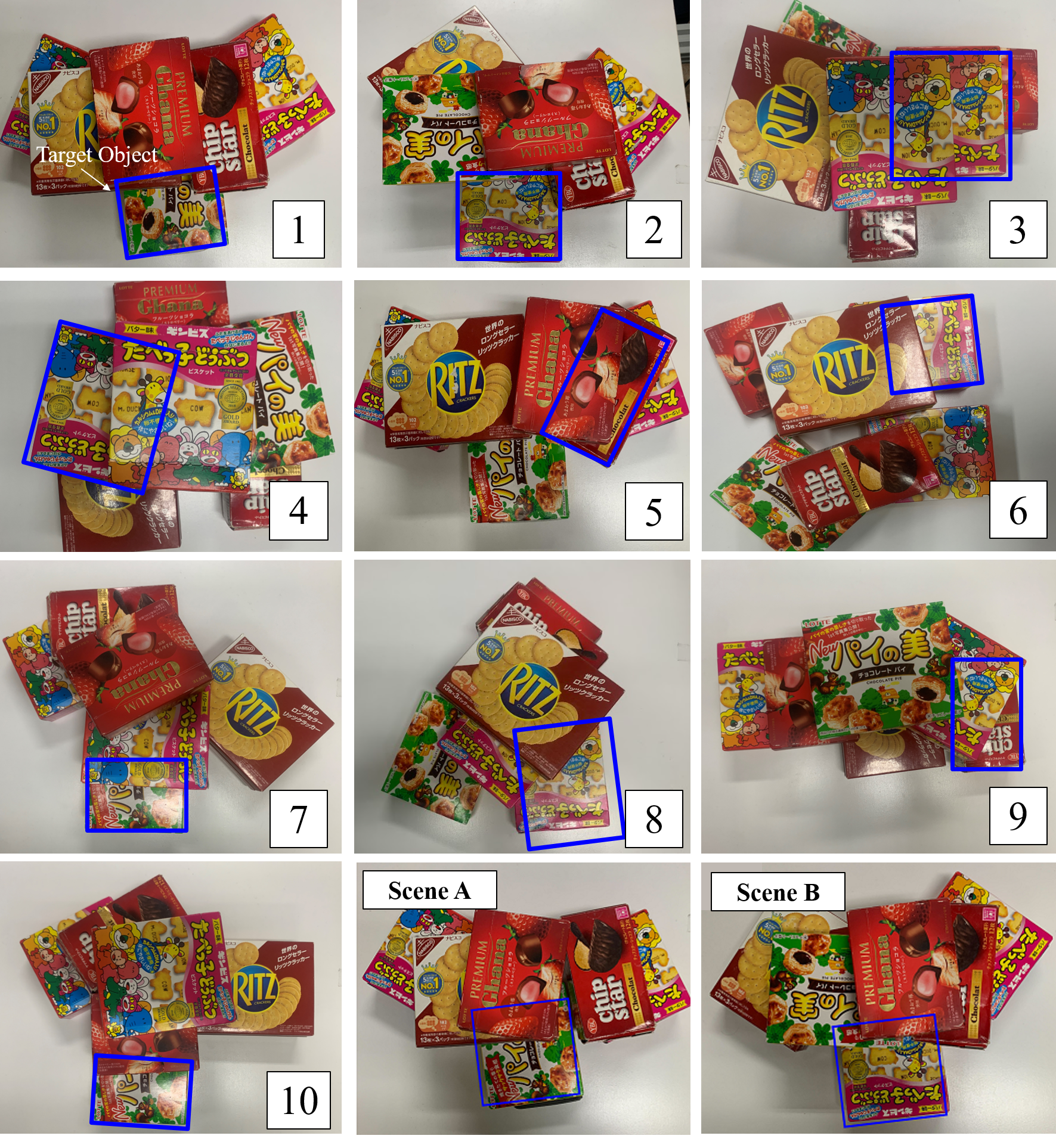}
            \caption{Piled objects used in our experiment where the target object is surrounded by the blue frame. (1-10) 10 patterns of piled objects. (Scene A,B) Additional 2 patterns of piled objects where we randomly changed the object poses. }
            \label{object}
        \end{center}
    \end{figure}
    
     We created 10 patterns of object piles as shown in Fig.~\ref{object}. The target object in the patterns 1,~2,~6--10 is placed on the table. In other cases, the target object is stacked on other objects. We experimented to retrieve the target object for these patterns of object piles. In addition, we added 10 additional experiments for each of the patterns 1 and 2 by modifying the objects' pose as shown in the scene A and B of Fig.~\ref{object}, where the amount of modification was randomly determined as far as the clutter does not collapse.
    
    The performance of the proposed method and other baseline algorithms are compared by computing the success rate. The success rate is evaluated by counting the trials in which the target object was grasped without collapsing the clutter before the final grasping action. We judged that the extraction was successful even if the clutter collapsed at the final grasping action.  
    The baseline methods used in this paper are:
    \begin{itemize}
        \item \textbf{Simple algorithm}, which is the heuristic-based method where the robot sequentially picks the objects included in the clutter from the top. The main purpose of this comparison is  to verify the optimality and capability of the safe action selection of our algorithm.
        \item \textbf{Single-arm algorithm}, which is the method in which the robot grasps the target object after sliding the target object without the support of the other hand. The purpose of this comparison is to verify the effectiveness of the bimanual manipulation. 
    \end{itemize}
    
    \begin{table}[tb]
     \caption{Comparison of other methods}
     \label{table}
     \centering
      \begin{tabular}{lccc}
       \hline
        & Num.~(\%) 
        & \begin{tabular}{c} Action \\ steps \end{tabular}
        & \begin{tabular}{c} Moving \\ Amount \end{tabular}~(mm)
        \\
       \hline \hline
       \textbf{Proposed method} & 7/10~(70.0\%) & \textbf{2.7} & \textbf{65.6} \\
       Single-arm & 5/10~(50.0\%) &  --- &  129.7 \\
       Simple algorithm & \textbf{8/10~(80.0\%)} & 3.1 & --- \\
       \hline
      \end{tabular}
    \end{table}

    \subsection {Results}

    \begin{figure*}[t]
        \centering
        \includegraphics[width=0.99\linewidth]{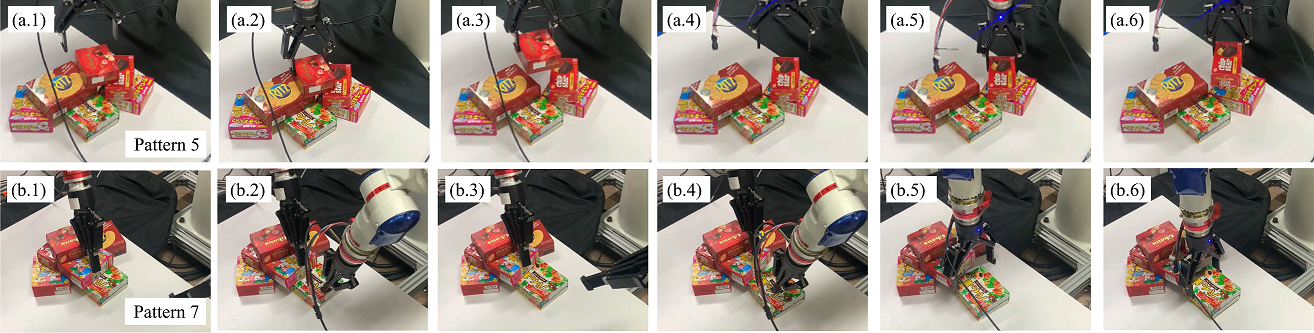}
        \caption{Experimental scenes. (a) The robot was able to grasp the target object after eliminating an obstacle (pattern 5 in Fig.~\ref{object}). 
        (b) First, the robot supported the object on the top of the clutter, and then slid the target object. Second, the robot supported another object and slid the target object again. In this case, the robot was able to select the appropriate action according to each scene (pattern 7 in Fig.~\ref{object}). }
        \label{exp}
    \end{figure*}
    
    The proposed method succeeded in extracting the target object from the pile 7 out of 10 times without causing the other objects 
     to fall ($70\%$). For the random arrangement of Scene A  and Scene B in Fig.~\ref{object}, the proposed method succeeded 6 out of 10 times for Scene A and 8 out of 10 times for Scene B.
    As seen in the experiment, when sliding the target object, one hand chose to slide it in the direction to reduce its occluded area while the other hand supports the object placed on top of the target one. When the removing action was judged to be feasible and efficient, the object selected the action to remove the object placed on top of the target one. Fig.~\ref{exp} shows the experimental scenes. 

    The failed cases include the target object failing to be grasped and the pile collapsing during the sliding action. For the second case, when there are no other objects in the opposite direction of sliding in the initial state, the object on the target object cannot be supported by other objects after sliding the target object. Therefore, the object on the target object will fall, which will cause the pile to collapse. 
    
  \begin{table}[tb]
     \caption{Results of two scenes with the proposed method}
     \label{table2}
     \centering
      \begin{tabular}{lccc}
       \hline
        Environment & Num.~(\%) 
        & \begin{tabular}{c} Action \\ steps \end{tabular}
        & \begin{tabular}{c} Moving \\ Amount \end{tabular}~(mm)
        \\
       \hline \hline
       Scene A & 6/10~(60.0\%) & 2.8 & 53.5 \\
       Scene B & 8/10~(80.0\%) & 2.3 & 77.6 \\
       \hline
      \end{tabular}
    \end{table}
    \subsection {Discussion}
    As shown in Table~\ref{table}, the proposed method showed improvements in both the action steps and the total amount of object movement. The success rate of the single-arm approach without dual arms was low because the object placed on the arm moved simultaneously during retrieval, making it difficult to detect the grasping posture of the object. The proposed method reduced the total amount of object movement. Even in situations where single-arm retrieval was challenging, the proposed method facilitated the removal of the object without disturbing the surrounding clutter. Thus, the proposed method is considered effective. 

    Although the success rate of the simple algorithm is slightly higher than that of the proposed method, as shown in Table~\ref{table}, it cannot be determined from the observation results whether an object in the middle of the pile can be grasped because the Simple algorithm tends to grasp the object at the top of the pile even if the target object is at the bottom of the pile. Accordingly, the success rate of the Simple algorithm may be lower in denser situations. On the other hand, the proposed method retrieves objects even in denser situations by extracting the object through action planning considering multiple possibilities. Moreover, the average number of movements of the proposed method is lower, indicating that it retrieves the object efficiently. In particular, the number of movements is significantly reduced when multiple objects overlap with the target object. 

    As shown in Table~\ref{table2}, Scene B has a higher success rate than Scene A where object positions are moved randomly. This is due to the initial placement of objects. Scene A has objects placed more horizontally than scene B. When objects are placed horizontally, objects placed on top are not supported by other objects and will fall when objects are pulled out. However, the object did not fall on the target object because object falling often occurs downward, and the object was successfully grasped in three of the four failed attempts.

    The reasons of failed object-picking are as follows:
    \begin{itemize}
        \item During the extraction process, only the object to be supported and the direction to be extracted are determined, but there are cases where the stack inevitably collapses. 
        \item It is difficult to accurately predict state transitions from the viewpoint directly above because the contact between objects cannot be correctly determined. 
        \item The gripper and the object collided due to the misalignment of the hand position caused by an observation error, and the gripping operation failed because the object size was too large in relation to the gripper opening width. 
    \end{itemize}
    Future work includes planning to reposition the supported object, modifying the state transition prediction based on the results of the first extraction and the prediction results, and planning actions based on more accurate predictions.

\section{CONCLUSIONS}
\label{sec:conclusion}
    In this work, we solve the problem of picking objects in a densely packed warehouse. 
    Our bimanual manipulation planner has achieved to slide the target object out of the clutter while supporting another object to prevent the clutter from collapsing. 
    Our method can predict stacking states from the observed clutter scene. It follows the Monte Carlo tree search to generate the efficient action. 
    Experimental results show that our proposed method reduces the number of motion sequences and prevents the clutter from collapsing with safe extraction of the target object. 
    Our future work will extend our framework in the following points: (1) extend the scope of application with other different shape objects in the experiments; (2) improve the transition model to apply object poses with a large pose tilt in clutter. 


\addtolength{\textheight}{-0cm}

\end{document}